\documentclass[letterpapaper, 10pt, journal, twoside]{IEEEtran}

\usepackage{amsmath,amsfonts}
\usepackage{array}
\usepackage[caption=false,font=normalsize,labelfont=sf,textfont=sf]{subfig}
\usepackage{textcomp}
\usepackage{stfloats}
\usepackage{url}
\usepackage{booktabs}
\usepackage{verbatim}
\usepackage{graphicx}
\usepackage{cite}
\usepackage[compact]{titlesec}
\usepackage{times} 
\usepackage[ruled,vlined]{algorithm2e}

\SetCommentSty{mycommfont}

\usepackage{newfloat}
\usepackage{listings}
\DeclareCaptionStyle{ruled}{labelfont=normalfont,labelsep=colon,strut=off} 
\usepackage{makecell}
\usepackage{multirow}
\usepackage{array}

\usepackage{cite}
\usepackage{amsmath,amssymb,amsfonts}
\usepackage[noend]{algpseudocode}
\usepackage[inline]{enumitem}
\usepackage{textcomp}
\usepackage{xcolor}
\usepackage{tikz}
\usetikzlibrary{positioning, shapes, fit, arrows}
\usepackage[cachedir=_minted,frozencache]{minted}

\usepackage{todonotes}
\usepackage{hyperref}

\makeatletter
\newcommand{\writings}[1]{%
    #1%
    \@ifnextchar.{}{%
        \@ifnextchar,{}{%
            \@ifnextchar'{}{%
                \@ifnextchar:{}{%
                    \@ifnextchar-{}{%
                    \@ifnextchar){}{\ }%
                    }%
                }%
            }%
        }%
    }%
}

\newcommand{\kb}[0]{\protect\writings{KB}}
\newcommand{\kbs}[0]{\protect\writings{KBs}}
\newcommand{\kbase}[0]{\protect\writings{knowledge-base}}
\newcommand{\Kbase}[0]{\protect\writings{Knowledge-base}}

\newcommand{\uc}[0]{\protect\writings{use-case}}

\newcommand{\Ucs}[0]{\protect\writings{Use-cases}}
\newcommand{\ucs}[0]{\protect\writings{use-cases}}
\newcommand{\storm}[0]{\protect\writings{Storm}}
\newcommand{\prism}[0]{\protect\writings{PRISM}}
\newcommand{\bw}[0]{\protect\writings{blocks-world}}

\renewcommand{\subsubsection}[1]{\noindent\textbf{#1}}

\makeatother
\newcommand{\pre}[1]{pre(#1)}
\newcommand{\eff}[1]{eff(#1)}

\definecolor{prismgreen}{rgb}{0, 0.6, 0}

\lstdefinelanguage{Prism}{ 
basicstyle=\color{red}\tiny\ttfamily, 
courier)
keywords=
{bool,C,ceil,const,ctmc,double,dtmc,endinit,endmodule,endrewards,endsystem,F,false,floor,formula,G,global,I,init,int,label,max,mdp,min,module,nondeterministic,P,Pmin,Pmax,prob,probabilistic,R,rate,rewards,Rmin,Rmax,S,stochastic,system,true,U,X},
keywordstyle={\bfseries\color{black}},
numberstyle=\tiny\color{black},
comment=[l] {//}, morecomment=[s]{/*}{*/}, 
commentstyle= \color{prismgreen}, 
tabsize=4, 
captionpos=b, 
escapechar=@ 
}

\newcommand{\cm}[0]{\textbf{\checkmark}}

\newcommand{\review}[1]{\textcolor{black}{#1}}

\newcommand{\savemargins}[0]{\vspace{0.2em}}

\setminted{baselinestretch=0.9}

\begin{document}

\title{Automated Generation of MDPs Using Logic Programming and LLMs for Robotic Applications}

\author{
    Enrico Saccon$^{1\star\dagger}$,~\IEEEmembership{Student Member, IEEE,}
    Davide De Martini$^{1\star}$, 
    Matteo Saveriano$^{2}$,~\IEEEmembership{Senior Member,~IEEE,}\\
    Edoardo Lamon$^{1}$,~\IEEEmembership{Member,~IEEE,} 
    Luigi Palopoli$^{1}$,~\IEEEmembership{Senior Member,~IEEE}
    Marco Roveri$^{1}$,~\IEEEmembership{Member,~IEEE,} 
\thanks{Manuscript received June 24, 2025; revised October 11, 2025.}
\thanks{This paper was recommended for publication by Editor Angelika Peer upon evaluation of the Associate Editor and Reviewers' comments. This work was co-funded by the European Union under NextGenerationEU (FAIR- Future AI Research - PE00000013), under project INVERSE (Grant Agreement No. 101136067), under project MAGICIAN (Grant Agreement n. 101120731) and by the project MUR PRIN 2020 (RIPER - Resilient AI-Based Self-Programming and Strategic Reasoning - CUP E63C22000400001). The data collection and experiments were conducted with the approval of the ethics committee at University of Trento under application No. 2025-003.}
\thanks{$^{1}$ Enrico Saccon, Edoardo Lamon, Luigi Palopoli and Marco Roveri are with Department of Information Engineering and Computer Science, University of Trento, Italy. {\tt\footnotesize name.surname@unitn.it}}
\thanks{$^{2}$ Davide De Martini and Matteo Saveriano are with the Department of Industrial Engineering, University of Trento, Italy. {\tt\footnotesize name.surname@unitn.it}}
\thanks{$^{\star}$ These authors contributed equally.} 
\thanks{$^{\dagger}$ Corresponding author: {\tt\footnotesize enrico.saccon@unitn.it.}}
\thanks{\textcopyright 2025 IEEE.  Personal use of this material is permitted.  Permission from IEEE must be obtained for all other uses, in any current or future media, including reprinting/republishing this material for advertising or promotional purposes, creating new collective works, for resale or redistribution to servers or lists, or reuse of any copyrighted component of this work in other works.}
}

\markboth{IEEE Robotics and Automation Letters. Preprint Version. Accepted November, 2025}{Saccon \MakeLowercase{\textit{et al.}}: Automated Generation of MDPs Using Logic Programming and LLMs for Robotic Applications}



\maketitle

\begin{abstract}
We present a novel framework that integrates Large Language Models (LLMs) with automated planning and formal verification to streamline the creation and use of Markov Decision Processes (MDP). Our system leverages LLMs to extract structured knowledge in the form of a Prolog knowledge base from natural language (NL) descriptions. It then automatically constructs an MDP through reachability analysis, and synthesises optimal policies using the Storm model checker. The resulting policy is exported as a state-action table for execution. We validate the framework in \review{three} human-robot interaction scenarios, demonstrating its ability to produce executable policies with minimal manual effort. This work highlights the potential of combining language models with formal methods to enable more accessible and scalable probabilistic planning in robotics.
\end{abstract}


This is the accepted version of the article published in IEEE Robotics and Automation Letters (2025). The final published version will be available via IEEE Xplore.



\section{Introduction}
\label{sec:introduction}

\IEEEPARstart{T}{he} most advanced frontier of robotics is the creation of machines that operate in uncontrolled environments and react to unanticipated conditions within acceptable safety margins. Human-robot interaction falls into this area. 
Humans can be unpredictable, either because they are not entirely focussed on the task or because they exercise their free will to change their behaviour in ways they consider more convenient. Fortunately, there are situations in which, even though humans can decide freely, their decisions are heavily influenced by social conventions. For example, in a collaborative construction task involving two humans, the set of available actions is relatively limited, and in most cases, one action is clearly preferable to the others. Structured behavioural patterns such as this are frequently modelled using probabilistic techniques.

For a robot to operate proficiently in such scenarios, it is essential to adopt a planning framework that defines policies, i.e., closed-loop schemes that associate actions to the perceived state of the system. The definition of the final goal can be complex to specify. For instance, in collaborative robotics we may wish to achieve the objective in a reduced time, leaving \emph{at the same time} the human co-worker with the freedom to choose their favourite course of actions. Our objective in this paper is to derive optimal policies of this kind from an informal specification in Natural Language (NL).

\noindent
\textbf{Related Work.}
\review{Markov Decision Processes (MDPs) are commonly used for modeling probabilistic decision making with efficient algorithms for policy generation under full and partial observability~\cite{kaelbling1998planning}. In robotics, MDPs have been applied in various areas such as decision support systems~\cite{DOLTSINIS2020103190}, motion planning~\cite{kurniawati2011motion}, optimal control~\cite{Xuchu2014Control}, planning under uncertainty~\cite{gopalan2017planning}, and coordinated multi-agent behaviour~\cite{bogert2014interacting}, including human adaptation~\cite{nikolaidis2017game} and preference inference~\cite{fern2010modeling}. As a result, MDPs are widely used in human-robot interaction~\cite{karami2011decision}.}
A key challenge in deploying MDPs is creating scalable models for real-time policy generation. This is difficult because human behaviour and dynamic environments are often informally described through language, images, or video, while MDPs require symbolic representations for classical computation.

Reinforcement Learning (RL)~\cite{kaelbling1996reinforcement} offers an alternative, as it learns policies through exploration without requiring prior domain specification. This flexibility has made RL popular in robotics~\cite{singh2022reinforcement}.
RL often faces sample inefficiency, reward design challenges, and high computational costs. Incorporating domain knowledge can help, though it's rarely easily obtained from informal sources. Large Language Models (LLMs) offer a promising solution. Initially developed for NL understanding~\cite{NEURIPS2023_b6b5f50a}, LLMs are now used for planning~\cite{ aghzal2025surveylargelanguagemodels} and robotics applications with uncertainty~\cite{mon_williams_embodied_2025}, showing excellent results, especially with few-shot learning, which uses examples to guide task performance~\cite{NEURIPS2020_1457c0d6}.
Several authors propose LLMs as standalone planners~\cite{ahn2022code, brohan2023rt2}, but they lack predictable, explainable behaviour, especially for complex environments where failure recovery is vital. To enhance reliability, researchers use LLMs to generate explainable representations like PDDL~\cite{oswald2024llms}, a formal language for planning problems. Extensions such as PPDDL~\cite{Younes2004PPDDL1} and RDDL~\cite{Sanner2011RDDL} enable probabilistic reasoning with MDP solvers. However, these languages have limited expressiveness and cannot fully capture the complexity of real-world robotic environments~\cite{7563060}.

Ontologies and \kbs effectively capture human knowledge and common sense. Their automated construction uses tools such as OLAM~\cite{lamanna2021online} and interactive learning~\cite{meli2021inductive}, although these rely on existing plans, lack generalisation, and do not model probabilistic aspects. LLMs offer a scalable solution, synthesising \kbs for various scenarios~\cite{xu2024large}.

Logic-based knowledge representations have been used in robotics for planning and automated reasoning~\cite{meli2023logic}. Probabilistic Logic Programming (PLP)~\cite{riguzzi2018survey} adds uncertainty to these representations. For example, distributional clauses~\cite{daan2015problog} facilitate the creation of a probabilistic Monte Carlo planner~\cite{nitti2017planning}. However, logic knowledge bases require a complete domain specification. Similarly,  the inference mechanism of ProbLog addresses MDPs with value iteration~\cite{bueno2016markov}, but also requires a fully specified domain.
In~\cite{saccon2024prolog}, LLMs are combined with the explainability and reasoning capabilities of logic programming, but it only supports deterministic planning, lacking the probabilistic aspect of human–robot interaction.

\noindent
\textbf{Paper Contribution}. The objective of this paper is to develop an open-source framework\footnote{\href{\detokenize{https://www.github.com/idra-lab/prolog_mdp}}{https://www.github.com/idra-lab/prolog\_mdp}} for generating probabilistic policies from an informal narrative that describes a robotic scenario. 
In this work, a scenario refers to a specific scene (e.g., an intersection involving humans and autonomous agents) that we describe in NL and utilise as part of the input.
The high-level textual narratives that describe the process or scenario and its desired behaviours are provided by domain experts interacting with the front end of the system, unaware of the underlying system functionality.
The system constructs, on the back end, a logical \kbase (\kb) that encodes entities and actions as predicates.
Such \kbs are human-readable by system designers and support inference of new facts using logical connectives. The role of system designers is to verify the correctness of the generated KB and tailor the LLM output to the specific problem through methods such as few-shot prompting or fine-tuning.
Inspired by PLP, our method links probabilities to predicates to automatically construct an MDP, which a state-of-the-art solver processes to generate a policy.
Our framework is designed to augment, not replace, human expertise: domain experts provide high-level conceptual narratives, while system designers ensure formal correctness. The system’s interpretable knowledge base enhances productivity, transparency, and trust by allowing experts to inspect and verify its decisions.
We demonstrate the efficacy of the framework in \review{three} use cases, chosen as representative paradigms of a broader class of applications in which robots operate in uncertain environments.  The results show that: 
\begin{enumerate*}[label=\roman*)]
\item a \kb can be efficiently derived from NL text, and 
\item the resulting MDP yields effective policies suitable for industrial applications.
\end{enumerate*}

\begin{figure*}[htp]
    \centering
    \includegraphics[width=\textwidth]{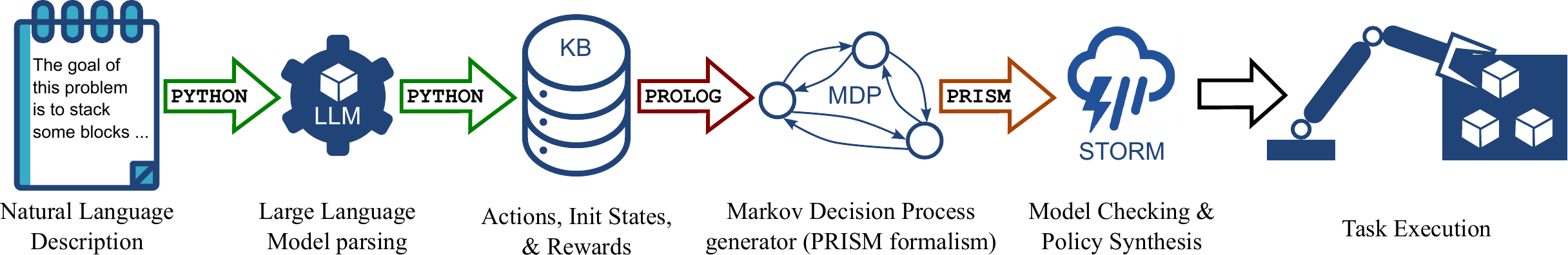}
    \caption{General diagram of the framework. Given an NL description of the query, it generates a Prolog \kb through few-shot prompting with an LLM. The \kb is used to extract a \prism MDP, which \storm uses to synthesize an optima policy. The policy can then be used to execute actions.}
    \label{fig:framework-diagram}
\end{figure*} 
\section{Background}
\label{sec:background}

\noindent\textbf{Planning.} %
We take inspiration from the STRIPS classical planning problem~\cite{ghallab03}, which describes an action $a$ by its preconditions $\pre{a}$, which must be satisfied in the current state to be executable, and its effects $\eff{a}$, which encode the result of its execution. 
\review{Taking the lines of PPDDL~\cite{Younes2004PPDDL1}, we consider an action to have more than one effect $\eff{a}_i$ associated with a probability $P_i \in [0,1]$ of being applied, such that $\sum_i P_i = 1$. Assuming a finite set of objects, the predicates and the set of actions induce an MDP~\cite{reason:Puterman90a}.}

\noindent\textbf{MDPs and Policies.}
An MDP is a mathematical framework for modelling sequential decision-making problems in environments characterised by stochasticity. It can be described with a 4-tuple $\langle \mathcal{S}, s_0, \mathcal{A}, \mathcal{P}, \mathcal{R} \rangle$, where: 
$\mathcal{S}=\{s_1, \hdots, s_N\}$ is a finite set of states representing descriptions of the system;
$s_0 \in \mathcal{S}$ denotes the initial state:
$\mathcal{A}=\{a_1, \hdots, a_M\}$ is a finite set of actions that cause transitions between states;
$\mathcal{P}_a(s,s')$ (transition probability relation) describes the probability of moving from state $s$ to state $s'$ given action $a$;
$\mathcal{R}$ is a reward function $\mathcal{R}(s,a)$ that assigns a value for executing action $a$ in state $s$.
%
%
Given an MDP and a \review{probabilistic temporal} property, a policy can be synthesised, which is a mapping of states to actions such that the trajectories (sequences of states satisfying the transition probability relation) satisfy the property~\cite{forejt:hal-00648037}.

\textbf{Prolog.} \review{Prolog is a logic programming language, applied in fields like computational linguistics~\cite{prologLing} and AI and robotics~\cite{prologRob}. It uses predicates and logic-based rules for symbolic reasoning on simpler facts stored in a \kb.}

\section{Proposed Framework}
\label{sec:our_framework}
 Fig.~\ref{fig:framework-diagram} visually represents the different blocks comprising our framework, all of which are described in detail below.
 
\textbf{Knowledge-base.} In our framework, we use Prolog\footnote{SWI-Prolog~\url{https://www.swi-prolog.org/}.}
to formally specify a \kb that stores:
\begin{enumerate*}[label=\alph*)]
    \item The initial state $s_0$ of the MDP, defined as a list of predicates.
    \item Grounded predicates that characterise the scenario presented in the input query, i.e., immutable characteristics such as different types of agents, e.g., a robotic arm or a wheeled robot.
    \item The actions that can be executed to transition from one state to another.
    \item The reward function to score the transitions.
    \item Auxiliary functions to write the MDP in the \prism formalism~\cite{forejt:hal-00648037}.
\end{enumerate*}
The Prolog predicates for reward functions are divided into two categories: \review{necessary} and \review{sufficient}. The former must be satisfied for the transition to be valid; if even one of them fails, the reward is set to a fixed value to discourage the agent from selecting actions leading to undesired states. If all \review{necessary} conditions are satisfied, the final reward is computed by adding the contributions of the \review{sufficient} conditions that match the transition. \review{Thus, the reward function is used both to prevent invalid transitions and to promote desired behaviours. The reward functions are pure Prolog code that the LLM generates.} 

\begin{figure}
    \centering
    \savemargins
\begin{minted}[fontsize=\footnotesize]{prolog}
init_variable_string("p1 : int init 0;\n
    p2 : int init 0;\n p3 : int init 0;\n").
\end{minted}
    \caption{An example of an auxiliary function to generate the PRISM variables: there are three pillars, \texttt{P1}, \texttt{P2}, and \texttt{P3}, which are integers and can have a value between 0 and 4, plus the reward value which is defined as an integer.}
    \label{fig:auxFunctionExample}
\end{figure}

\noindent\textbf{\kb Generation.} 
\begin{figure}[t!]
    \centering
\begin{minted}[escapeinside=||,fontsize=\footnotesize,breaklines,breaksymbolleft=]{text}
The use-case consists of an industrial Autonomous Guided Vehicle (AGV) that has to carry out a task. The task is completed when the AGV has crossed the entire factory without emergency stops.

|\textcolor{olive}{\bfseries ... \emph{full text in the supplementary material}}|

The goal would be to cross the factory in the least amount of time possible.
\end{minted}
    \caption{\review{Example of a query passed as input to the LLM to generate the MDP.}}
    \label{fig:query_ex}
\end{figure}
\begin{figure}[t!]
    \centering
\begin{minted}[escapeinside=++,fontsize=\footnotesize,breaklines,breaksymbolleft=]{yaml}
Q:
 role: 'user'
 content: |
  How should I divide the code?
A:
 role: 'assistant'
 content: |
  When generating an answer, remember to use the correct Markdown tags, which are:
  - `kb` for the initial state;
  - `action` for the actions;
  - `reward` for the rewards.
  +\textcolor{olive}{\bfseries ... \emph{full text in the supplementary material}}+
\end{minted}
    \caption{An example of how the output from the LLM should be formatted.}
    \label{fig:general_ex}
\end{figure}
\begin{figure}[t!]
    \centering
\begin{minted}[escapeinside=++,fontsize=\footnotesize,breaklines,breaksymbolleft=]{yaml}
role: 'assistant'
content: |
 Sure. Given the description, we can define the initial state a composition of the following variables: Section: 1, EStop: false, Delay: 0. 
 The initial state can be represented as:
 ```kb
 % Initial state
 initial_state([section(0), estop(false), delay(0)]). ```
 +\textcolor{olive}{\bfseries ... \emph{full text in the supplementary material}}+
\end{minted}
    \caption{An answer used as example for the LLM, containing how to generate the \kb from the query shown in Fig.~\ref{fig:query_ex}.}
    \label{fig:answer_kb_ex}
\end{figure}
\begin{figure}[t!]
    \centering
\begin{minted}[escapeinside=++,fontsize=\footnotesize,breaklines,breaksymbolleft=]{yaml}
role: 'assistant'
content: |
 Sure. The actions that we need to perform are two: one for waiting and one for proceeding. ...
 +\textcolor{olive}{\bfseries ...}+
 The resulting actions will be: 
 ```actions
 % Action: wait
 % Preconditions: AGV in section (1 to 4), no estop
 % Effects: delay plus 20, change section, no estop
 action(wait(Section, Delay), ```
 +\textcolor{olive}{\bfseries ... \emph{full text in the supplementary material}}+
\end{minted}
    \caption{An answer used as example for the LLM, containing how to generate the actions. The query contains both the \kb (Fig.~\ref{fig:answer_kb_ex}) and the NL text (Fig.~\ref{fig:query_ex}).}
    \label{fig:answer_ac_ex}
\end{figure}
\begin{figure}
    \centering
    \savemargins
    \begin{minted}[escapeinside=++,fontsize=\footnotesize,breaklines,breaksymbolleft=]{yaml}
role: 'assistant'
content: |
 Sure. The rewards for the AGV use-case can be defined based on the provided description. From the description, there are two type of rewards:
 - the necessary one is that the AGV will not crash with the workers, and 
 - the sufficient ones are that the delay is minimized and that the task is completed. 
 % Sufficient condition: complete task
 suff_condition(complete_task, _State, [Section, _, _], _Action, Reward) :-
    Section = 5, Reward is 100. ```
    +\textcolor{olive}{\bfseries ... \emph{full text in the supplementary material}}+
\end{minted}
    \caption{An answer used as example for the LLM, containing how to generate the rewards functions from the NL text. The query for this answer contains the previously generated \kb, actions and the NL description shown in Fig.~\ref{fig:query_ex}.}
    \label{fig:answer_rew_ex}
\end{figure}
This step is performed using an LLM via few-shot prompting (Sec.~\ref{sec:background}). The LLM takes as input both the NL description of the domain elements (Fig.~\ref{fig:query_ex}), and a set of curated examples, containing some general examples (Fig.~\ref{fig:general_ex}) and others that are use-case-specific (Fig.~\ref{fig:answer_kb_ex}~to~\ref{fig:answer_rew_ex}). 
The former are used to give a high-level description of how the LLM has to generate the \kb, including both how the desired output should be structured and some focal rules that will be used to correctly parse the NL description in order to generate a correct and coherent \kb. The Python parser requires the LLM output to contain Markdown-like tags that allow for easily parsing the output and capturing the required information. The tags are  \verb|kb| (Fig.~\ref{fig:answer_kb_ex}), \verb|actions| (Fig.~\ref{fig:answer_ac_ex}), and \verb|rewards| (Fig.~\ref{fig:answer_rew_ex}). 

The general examples describe technical aspects, for example, how an action should be structured, as in using the predicate \verb|action| with arguments the name with the action arguments, and the lists of preconditions and effects, how the initial state should be formatted, i.e., as a list contained in the predicate \verb|initial_state|, \review{and the labels for \storm, since we want to either maximize the probability of success of a label or minimize the reward value}. 

The use-case-specific examples define how the NL should be converted into a \kb for that specific task. To provide the LLM with this information, we break down the query and examples into different parts, each corresponding to the components of an MDP: the actions, the reward, and the states. The examples explain both how the \kb shall be formatted (see Fig.~\ref{fig:answer_kb_ex}~to~\ref{fig:answer_rew_ex}), and also include counterexamples that illustrate wrong modelling choices. In Fig.~\ref{fig:answer_kb_ex}, we see that the  LLM shall generate the initial state in the \prism formalism, as well as the other predicates, e.g., \verb|get_state| and \verb|get_printable_state| that are problem dependent and hence must change depending on the input query.  

The LLM generates all the components of the \kb: initial state, grounded predicates, actions, reward functions, and auxiliary functions to generate the \prism file. These parts of the \kb are produced in multiple steps and, to ensure consistency in the terminology and structure of the predicates, the framework feeds back the previously generated information into the prompt at each subsequent step. We adopted this incremental approach rather than generating the entire \kb in a single pass, as preliminary experiments indicated that it would provide more accurate and coherent outputs from the LLM. For example, we begin with the query presented in Fig.~\ref{fig:query_ex}, which is then passed to the LLM along with a series of examples. The LLM returns a well-formed output containing the \kb within the tags \verb|kb|, which is then parsed and fed-back to the LLM with the same original query, and we ask the LLM to generate the actions. As in the previous step, we extract the relevant part for the actions from the output and then ask the LLM to generate the final part for the rewards, feeding both the \kb and the actions, plus the original query. 

\review{Importantly, the reward functions are \emph{automatically generated by the LLM}. Drawing on its internal knowledge and the few-shot examples, the model not only infers the appropriate reward structure for the given scenario but also distinguishes between \emph{necessary} and \emph{sufficient} conditions. As mentioned before, the formers correspond to constraints not to be violated, whereas the latter correspond to desirable but non-mandatory objectives. This classification is entirely handled by the model during the reward generation step, based on the patterns and instructions provided in the examples and its internal logic.}

Finally, a human operator is in charge of checking the quality of the produced \kb and eventually fixing the mistakes.

\begin{figure}[t!]
\savemargins
    \centering
    \scalebox{0.7}{\begin{tikzpicture}[
    scale=1, 
    fatnode/.style={circle, fill=blue!20, minimum size=5mm},
    actionnode/.style={circle, fill=green!20, minimum size=5mm},
    fakenode/.style={circle, fill=white, minimum size=1mm}
    ]
    
    \node[fatnode] (1)      at (0,0)        {$s_1$};
    \node[actionnode] (2)   at (0, -1.3)    {$E_1$};
    \node[fatnode] (3)      at (-1, -2.5)   {$s_2$};
    \node[fatnode] (4)      at (1, -2.5)    {$s_3$};

    \node[fakenode] (5) [right=of 2] {};
    \node[fakenode] (6) [right=of 5] {};

    \draw[thick, ->] (1) -- (2) node[midway, left] {0.9};
    \draw[thick, ->] (2) -- (3);
    \draw[thick, ->] (2) -- (4);

    \draw[thick, ->] (5) -- (6);

    \node[fatnode] (7)      at (4.5, 0)       {$s_1$};
    \node[actionnode] (8)   at (4.5, -1.3)    {$E_1$};
    \node[fatnode] (9)      at (3.5, -2.5)    {$s_2$};
    \node[fatnode] (10)     at (5.5, -2.5)    {$s_3$};

    \draw[thick, ->] (7) -- (8);
    \draw[thick, ->] (8) -- (9) node[midway, above left] {0.45};
    \draw[thick, ->] (8) -- (10) node[midway, above right] {0.45};

\end{tikzpicture}}
	\caption{Example of how to distribute the probability: assume that $s_i$ are the states, $E_1$ is an effect of the action with an associated probability of 0.9. Assuming a uniform distribution, the states $ s_2$ and $ s_3$, generated by applying the same lifted effects $E_1$, have the same probability of being reached; hence, the probability 0.9 is split between the two, as shown on the right.
    }
    \label{fig:prob_distr}
\end{figure}
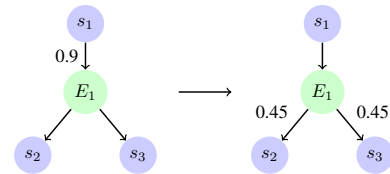

\noindent\textbf{MDP Generator.} 
This module, implemented in Prolog, generates the MDP from the \kb and the specific query with the following four steps.
{\it Step 1: Graph generation. }
The MDP generator module computes a graph $G = \langle S, E\rangle$ starting from an initial state corresponding to the initial condition. 
The set of nodes $S$ comprises all possible states of the MDP, and the edges $E$ correspond to the transitions: each transition (edge) is associated with an action $a$, with the probabilistic effects $E_i\in\eff{a}$ applied during the transition, and with the corresponding probability $P_i$.
The algorithm recursively checks if an action $a$ from the action space $\mathcal{A}$ can be executed. If, in the current state $s$, the action's preconditions $\pre{a}$ are satisfied, then the probabilistic effects are applied, enabling the algorithm to generate new states. This is repeated by recursively considering each set $E_i$ in the probabilistic effects $\eff{a}$ and by checking that the delete effects within $E_i$ have grounded arguments. If they have, the effects are grounded and applied, generating one or more new states. It is important to note that a single set of effects $E_i$ can transition to multiple states as the values of the lifted predicates are recursively and exhaustively assigned. For example, let $s_c$ be the current state in which it is possible to apply the action $a$=\verb|move_block| with the set of effects $E_i={add(block(B))}$, and let the \kb contain the predicates \verb|block(b1), block(b2)|. The action $a$ fromstate $s_c$ leads to two different states $s_{n1}, s_{n2}$, one containing \verb|block(b1)| and the other containing \verb|block(b2)|. This is done by exhaustively grounding all the lifted predicates in the lists of effects of the action. 
If any of the delete effects of $E_i$ cannot be grounded in the current state, this means that $E_i$ cannot be applied in the current state (we expect them to hold in the current state; otherwise, there is no reason to delete them). However, unlike preconditions that prevent an action from being applied, here we add a self-transition $edge_{\langle a, E_i\rangle}(s,s, P)$ to $G$. This interprets the case as a no-op (do nothing).
Upon the generation of a new state $s'$, the algorithm evaluates whether it is already included within  $S$ of $G$. If included, the algorithm merely adds a new transition from $s$ to $s'$. Conversely, if the state is absent, the new state $s'$ is incorporated into $S$ prior to the addition of the transition. 
The function is then invoked recursively on it ensuring a complete reachability analysis, which, therefore, ends when every possible combination of actions, states, and grounding possibilities has been examined.
\begin{algorithm}[t!]
\small
\SetAlgoNlRelativeSize{-1}
\SetKwFunction{generateMDP}{generate\_mdp}
\SetKwFunction{applyEffects}{apply\_effects}
\SetKwFunction{selectAction}{select\_action}
\SetKwFunction{checkPreconditions}{check\_preconditions}
\SetKwFunction{addNode}{add\_state}
\SetKwFunction{addEdge}{add\_edge}
\SetKwFunction{ground}{ground}
\SetKwFor{ForEach}{for}{do}{endfor}
\SetKwFor{While}{while}{do}{endwhile}
\SetKwInOut{Input}{Input}
\SetKwInOut{Output}{Output}
\SetKwProg{Fn}{Function}{:}{}
\SetInd{0.5em}{1em}

\Fn{\generateMDP{$s$}}{
    \While{$a \gets$ \selectAction{$a$}}{
        \checkPreconditions{$a$}\;
        \ForEach{$(Prob, E_i) \in \eff{a}$}{
            \applyEffects{$s, s', a, E_i, Prob$}\;
        }
    }
}

\Fn{\applyEffects{$s, s', a, E_i, Prob$}}{
    \ForEach{del(Predicate) in $E_i$}{
        \If{\ground{Predicate}}{
            \If{$s'$ is $\neg$Visited}{
                \addNode{$s'$}\;
                \generateMDP{$s'$}\;
            }
            \addEdge{edge($s, s', a, E_i, Prob$)}\;
        }
        \Else{
            \addEdge{edge($s, s, a, E_i, Prob$)}\;
        }
    }
}
\caption{The MDP generation functions}
\label{algo:empty_state}
\end{algorithm}
{\it Step 2: Generation of the Transition Probabilities.}
After $G$ has been generated, the next step is to distribute the probabilities for its transitions. The probabilities $P_1, \ldots, P_n$ associated in \kb with an action's set of effects, $\eff{a}=\{E_1, \ldots, E_n\}$, do not depend on the number of states generated by applying that action's effects. This means that each probability $P_i$ must be distributed over the possible transitions created by the action.
To perform this task, the framework does a complete traversal of $G$ and, for each edge, if 
\begin{enumerate*}[label=\arabic*)]
    \item it has the same starting state,
    \item it is associated with the same action $a$, and
    \item it originates from the same set of effects $E_i$, but
    \item leads to different new states,
\end{enumerate*}
then its probability $P_i$ is distributed across the created transitions. For an illustration, see Fig.~\ref{fig:prob_distr} and Algorithm~\ref{algo:prob_distr}. This step generates the graph $G'$, which is associated with $G$ and encodes the transition probabilities.
For instance, consider the following toy example: the current state has predicate \texttt{position(1, 0), position(2,0)}, meaning that there are two pillars that have no blocks. If we consider the action \texttt{bi}, meaning that we propose a block of type base and one of type intermediate on the tray, with probabilities 0.75 and 0.25 of being taken by the user, respectively, then the effect \texttt{0.75:[del(position(Pillar, 0)), add(position(Pillar, 1))]} corresponding to the user choosing the block of type base, has two possible outcomes: \texttt{Pillar=\{1, 2\}}, each with probability $0.75/2$, and each leading to a different state.

\begin{algorithm}[t!]
\small
\SetAlgoNlRelativeSize{-1}
\SetKwFunction{refineProbabilities}{refine\_probabilities}
\SetKwFunction{findall}{findall}
\SetKwFunction{changeProb}{change\_prob}
\SetKwInOut{Input}{Input}
\SetKwInOut{Output}{Output}
\SetKwProg{Fn}{Function}{:}{}
\SetInd{0.5em}{1em}

\Fn{\refineProbabilities{$G=(S, E)$}}{
\textbf{Initialize} changed\_edges $\leftarrow [\,]$\;
\ForEach{e in $E$}{
    \If{e $\notin$ changed\_edges}{
        $\mathcal{E} \gets$ \findall{edge(e.$s$, e.$s'$, e.$a$, e.$eff$, \_)}\;
        new\_prob $\gets$ e.$prob / |\mathcal{E}|$\;
        \changeProb{new\_prob, $\mathcal{E}$}\;
        changed\_edges.append($\mathcal{E}$)\;
    }
}
}
\caption{The transition probabilities refinement function}
\label{algo:prob_distr}
\end{algorithm}

{\it Step 3: Generation of the rewards. }
To assign reward values to the transitions of the MDP graph $G'$, we distinguish between necessary and sufficient rewards. Necessary rewards correspond to strict constraints that must always hold. If even one necessary condition is violated, the transition is immediately assigned a fixed large negative penalty, and no further conditions are evaluated. This ensures that behaviours violating mandatory requirements are strongly discouraged.

If all necessary conditions are satisfied, the system then evaluates the sufficient rewards. These represent desirable but non-mandatory properties and contribute non-negative values (i.e., positive or zero) that accumulate when conditions are met. The more sufficient conditions are satisfied, the higher the reward of the transition. This formulation allows us to separate hard constraints from soft preferences, enabling policies that are both correct and optimised for user-specified objectives.

For example, consider the \emph{structure building} scenario with the policy ``Build the pillars by layers, while maximizing the user’s freedom''. The requirement that no two pillars differ in height by more than one block constitutes a \emph{necessary reward}: any violation results in a severe negative penalty. Conversely, providing the user with blocks that differ from one another represents a \emph{sufficient reward}: its fulfillment yields a positive bonus, but its absence does not incur any penalty.

{\it Step 4: PRISM Script Compilation.}
The compilation of the graph $G'$ into a \prism script is performed in three steps.
First, the \prism header is created using the LLM-generated strings as auxiliary functions for variable initialisation (see Paragraph {\it \Kbase }in Section~\ref{sec:our_framework}). Second, the transition model is built by traversing the graph (as an MDP) and generating the transitions, formatted using the LLM-generated predicates. Finally, it reports the labels generated by the LLM from the the NL input, and the complete MDP is saved in a file for processing by \storm. As mentioned earlier, the LLM system generates two labels for the MDP: one to maximize the probability of success (\verb|doneP|) and another to minimize the reward (\verb|doneR|). This distinction exists because the optimal final state may differ depending on the objective. For example, in the \bw \uc, both labels coincide since the task requires all three pillars to be built. In contrast, the AGV \uc has different labels: \verb|doneP| requires reaching the final section without emergency stops, while \verb|doneR| focuses on reaching it as quickly as possible, regardless of stops. \review{When compiling Storm, it is up to the user to choose which label is more appropriate, depending on whether they want to prioritize the successful execution of the task (\texttt{doneP}), or whether they want to optimize a reward value \texttt{doneR}. The two labels are \storm functionalities, which will then be used in the model checker to find the optimal solution for the provided label.}

\noindent\textbf{Policy synthesis and execution.} 
A policy for a given property is synthesized using the \storm solver~\cite{hensel2020probabilisticmodelcheckerstorm}. \storm provides a Python wrapper for easy integration. To obtain a policy from \storm, specify an optimization goal as a property. The common property is \texttt{Pmax=? [ F "doneP" ]}, which asks \storm to find a policy maximizing the expected probability of reaching states where "\texttt{doneP}" holds. For reward optimization, minimizing the reward is achieved by setting the property to: \texttt{Rmin=? [ F "doneR" ]}. The synthesized policy is stored as a \emph{state-action table}, which can be implemented in any robotic runtime environment, such as the adopted ROS2.

\begin{figure}[t]
\savemargins
    \centering
    \includegraphics[width=0.6\linewidth]{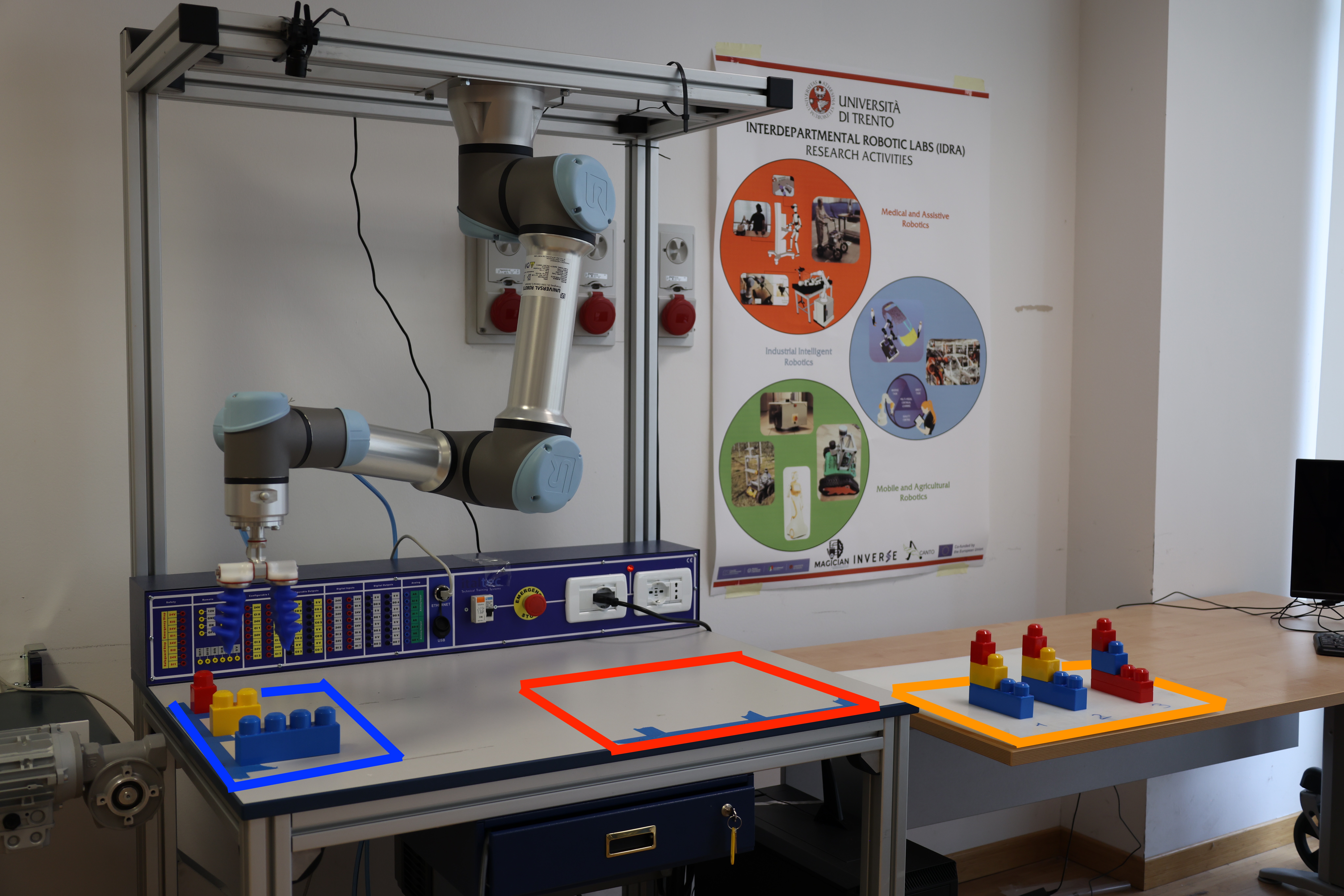}
    \caption{Experimental setup for the \bw scenario.} 
    \label{fig:lab-setting}
\end{figure}

\begin{figure*}[t!]
\savemargins
    \small
   \centering
\begin{minipage}{0.15\textwidth}
\begin{minted}[fontsize=\footnotesize]{prolog}
init_state([
    pil(1,0),
    pil(2,0),
    pil(3,0)]).
\end{minted}
\end{minipage}
\hfill
\begin{minipage}{0.83\textwidth}
\begin{minted}[fontsize=\footnotesize]{prolog}
action(tti(P1, P2),                            % action_name(args)
  [],                                          % AND Preconditions
  [pil(P1, 2), pil(P2, 1)],                    % OR preconditions
  [],                                          % Predicates to verify
  [0.9:[del(pil(P1,2)),add(pil(P1,3))], 0.1:[del(pil(P2,1)),add(pil(P2,2))]]). % Effects
\end{minted}
\end{minipage}
\vspace{0.5em}
    \caption{Initial state (left): no block has been placed. An example of actions (right) with the list of preconditions and effects.}
    \label{fig:KBuc}
\end{figure*}

\section{\Ucs and Experiments}
\label{sec:usecases}

In this section, we present \review{three} \ucs on which we tested the system. For each, we first outline the scenario, then provide a detailed explanation of the implementation and a description of the devised experiments.
We carried out five tests for each \uc to validate the framework's ability to generate a correct policy and to highlight the generalisation capabilities of the domain generation step.

\subsection{Structure Building}
Inspired by the \bw scenario~\cite{blocksworld} widely used in task planning, we devised a proof-of-concept scenario in which a human agent and a robotic manipulator cooperate to construct a structure. 
The \uc was tested both in simulation and in a real execution environment.

\noindent\review{\noindent\textbf{Problem description.} A human operator must build $N$ towers using different blocks (orange rectangle in Fig.~\ref{fig:lab-setting}). A robotic manipulator places the blocks on a tray (red rectangle) at predefined positions, all reachable but at varying distances from the human’s hands, hence with increasing probability of choosing a block based on proximity. The operator selects one block from the tray, discarding the others. The robot can draw any block from an infinite storage area (blue rectangle).}

\noindent \textbf{\kb.} 
\review{Fig.~\ref{fig:KBuc} shows an excerpt of the \kb for the scenario. Fig.~\ref{fig:KBuc} illustrates the initial state with no blocks on pillars. The right side displays an action placing two top blocks (\texttt{t}) and one intermediate block (\texttt{i}) on the tray, with the top blocks nearer to the human. An action includes a name with arguments, preconditions, and probabilistic effects. The \texttt{tti} action needs at least a pillar of height 2 or 1.}

\noindent \textbf{Reward value.} We want to encourage freedom of choice, hence the reward values are chosen so that the system encourages diversity of the blocks in the tray (as far each block corresponds to an admissible choice).

\begin{figure*}
    \small
\begin{minipage}{0.15\textwidth}
\begin{minted}[fontsize=\footnotesize]{prolog}
init_state([
  section(1),
  estop(0),
  delay(0)]).
\end{minted}
\end{minipage}
 \hfill
\begin{minipage}{0.9\textwidth}
\begin{minted}[fontsize=\footnotesize]{prolog}
action(proceed(S),                                      % action_name(args)
  [section(S), estop(0)],                               % AND preconditions
  [],                                                   % OR preconditions
  [Pno is 1-S/10, Pyes is S/10, NS is S+1, S<5]         % Predicates to verify 
  [Pyes:[del(estop(0)),add(estop(1))], Pno:[del(section(S)),add(section(NS))]]). % Effects
\end{minted}
\savemargins
\end{minipage}
    \caption{Initial state (left): AGV is at starting state without any emergency stop and delay. An example of actions (right) in which the predicates to verify plays a crucial role to also compute the probabilities of the different effects.}
    \label{fig:KBagv}
\end{figure*}

\noindent \textbf{Experiments.}
We performed five tests for this \uc:
\begin{enumerate*}[label=\arabic*)]
    \item There are a total of 3 types of blocks and the robotic arm places 3 blocks on the tray. The operator has to create 3 towers picking blocks from the provided ones. This case is the one we also used in the examples for the LLM.
    \item The probabilities of picking the blocks on the tray are changed with values not present in the examples.
    \item The number of pillars increases to five.
    \item The number of block types increases to four, and the number of pillars reduces to two.
    \item We considered two pillars, three types of blocks, and a new action that can be executed only at the end to place an architrave on top of the pillars.  
\end{enumerate*}
Tests are conducted to highlight the ability of the framework to generalise in the scenario and extract consistent policies.

\noindent \textbf{Real-world experiment.} Once the state-action table is extracted, it can be used in any ad-hoc system. We developed a ROS2 workspace to conduct real-world experiments.

\subsection{Traffic management of AGVs in an industrial scenario.} 

\noindent\textbf{Problem description. } 
\review{An Autonomous Guided Vehicle (AGV) must navigate a factory floor, each route is divided into sections and in each section there are workers who might disrupt its path. More workers increase the chance of emergency stops to prevent collisions. The objective is to traverse all sections without stops. The AGV has two actions: \texttt{wait}, which slows down to avoid crashes but delays the task, or \texttt{proceed}, which moves ahead without delay, risking an emergency stop. If the AGV stops, it ends negatively, as indefinite waiting causes delays that reduce factory efficiency.}

\noindent\textbf{\kb.} 
The initial state includes three predicates: tracking the section, reporting an emergency stop, and expressing delay (left of Fig.~\ref{fig:KBagv}). Two actions model this \uc by evaluating predicates to determine effect probabilities, incrementing the section, and increasing the delay.

\noindent\textbf{Reward \review{value}.} 
The goal is to reduce the likelihood of entering emergency stop mode while ensuring factory efficiency. Mandatory properties  in the NL text include human safety, while optional ones like minimizing time spent in a section help compute reward values. The reward function penalizes AGV emergency stops, and rewards for transitions without stops are inversely related to accumulated delay.

\subsubsection{Experiments.}
For this \uc, we rely on simulation to validate the proposed approach  since we have no access to the necessary hardware and resources.
We considered the following scenarios.
\begin{enumerate*}
    \item There are a total of 5 sections that the robot has travel across. The two only actions that it can do are: \verb|wait| and \verb|proceed|, as aforementioned.
    \item It extends the first test by increasing the section number to 10 sections in total. 
    \item It decreases the delay caused by the wait action by 15.
    \item It combines the second and third tests together, and also changes the \verb|wait| action. The robot now stops and it can only advance to the next section if \verb|proceed| is called. 
    \item A third previously unseen action, namely \verb|speed-up|, has to be generate by the LLM. This action increases the velocity of the AGV setting the delay to 0, but it also increases the probability of an emergency stop. \verb|proceed| now allows for moving to the next section with a reduced probability of collision.
\end{enumerate*}

\subsection{\review{Gripper domain}}
In the gripper domain~\cite{McDermott_2000}, a robot with two grippers moves balls between rooms. This domain tests our system's generalisation, serving as a balance between the first two \ucs by considering different settings with probabilistic actions, multi-stage transitions, resource constraints, and value prioritization.

\begin{enumerate*}[label=\arabic*)]
\item A robot with two grippers moves 3 balls from Room A to Room B. The move is certain, but picking succeeds at 0.9, and dropping at 0.95. \item Two labelled balls are involved: one is in Room A, the other is in Room B, requiring return to Room A before delivery. Movement succeeds at 0.98, picking at 0.88, and dropping at 0.94. 
\item In a three-room setup, the robot moves balls from Room A through a Corridor to Room B. Initial setup includes 2 balls in Room A and 2 in the Corridor. Move success is 0.97 from Room A to Corridor, and 0.95 from Corridor to Room B. Picking succeeds at 0.90, dropping at 0.93. 
\item Energy is limited: 4 fragile balls need delivery with 10 energy units. Each move, pick, or drop uses 1 unit of energy. Pick and drop success rates are 0.92 and 0.97. Task fails if energy depletes before completion. 
\item Delivery prioritization has 4 balls in Room A, with 2 as high-value. Movement is certain. High-value balls pick at 0.85, drop at 0.90; standard balls pick at 0.90, drop at 0.96. Carrying both high-value balls reduces drop success by 0.03.
\end{enumerate*}

\subsection{Implementation details and experimental setup.}
All experiments were executed on a desktop PC running Ubuntu 22.04 with an AMD Ryzen 7 7700X CPU and 64GB of DDR5 memory. We used Swipl version 9.2.9, and stormpy version 1.9.0.
For the generation of the \kb, we used GPT-4o \review{and GPT-5-mini} as the LLM. To reducestochasticity, we set the temperature to 0 and fixed the seed (to 42) beforehand. Although this choice does not produce repeatable behaviours, it significantly alleviates the unpredictability of the LLM.
\section{\review{Experimental Results}}
\label{sec:results}

\review{We analyse the results obtained in the three domains described in Sec.~\ref{sec:usecases}. 
Table~\ref{tab:res} presents results for the various steps of our framework, while Table~\ref{tab:policyRes} shows policy execution.}

\begin{table*}[ht]
\centering
\caption{Results of the experiments on the proposed \ucs averaged over 100 trials. In \sc{KB generation}, we report \cm if the LLM output was correct, or we use $X(N,M)$, where $N$ is the number of logical errors and $M$ the number of corrections. \sc{MDP extraction} shows the time to generate the MDP, refine probabilities (Sec.~\ref{sec:our_framework}), and write to file. \sc{Policy extraction} reports the average times (over 10000 runs) to extract the optimal policy for probability maximisation (top) and reward maximisation (bottom).}
{
\newcolumntype{C}{>{\centering\arraybackslash}p{0.6cm}}
\renewcommand{\arraystretch}{0.8}
\begin{tabular}{rCCCCCCCCCCCCCCC}
\toprule
                    & \multicolumn{5}{c}{\sc{Structure Building}}        & \multicolumn{5}{c}{\sc{AGV}}                      & \multicolumn{5}{c}{\sc{\review{Grippers}}}\\
                    & 1       & 2       & 3         & 4        & 5       & 1       & 2        & 3       & 4       & 5        & \review{1}       & \review{2}       & \review{3}        & \review{4}         & \review{5}         \\
\midrule
\sc{\# states}      & \sc{64} & \sc{64} & \sc{1024} & \sc{125} & \sc{17} & \sc{35} & \sc{120} & \sc{48} & \sc{78} & \sc{194} & \sc{14} & \sc{17} & \sc{104} & \sc{2027} & \sc{263} \\
\sc{\# actions}     & \sc{27} & \sc{27} & \sc{27}   & \sc{16}  & \sc{28} & \sc{2}  & \sc{2}   & \sc{2}  & \sc{2}  & \sc{3}   & \sc{59} & \sc{5}  & \sc{12}  & \sc{18}   & \sc{19}  \\
\midrule
\makecell{\sc{\review{\kb  GPT-4o}}} 
                    & \cm     & \cm     & \cm       & \cm      & \cm     & \cm     & \cm      & \cm     & \cm     & \cm      & \cm     & \cm     & \cm      & X(2,2)   & X(1,1) \\
\makecell{\sc{\review{\kb  GPT-5-mini}}} 
                    & \cm     & \cm     & \cm       & \cm      & \cm     & \cm     & \cm      & \cm     & \cm     & \cm      & \cm     & \cm     & \cm      & \cm   & \cm \\
                                
\midrule
\multirow{3}{*}{\sc{\makecell{MDP\\extraction}}}
                    & 0.073 & 0.074 &  28.928 & 0.055 & 0.006 & 0.001 & 0.005 & 0.001 & 0.003 & 0.014 & 0.008 & 0.002 & 0.021 & 3.184 & 0.080 \\
                    & 0.788 & 0.792 & 780.792 & 0.270 & 0.053 & 0.000 & 0.001 & 0.000 & 0.001 & 0.004 & 0.009 & 0.001 & 0.004 & 0.999 & 0.070 \\
                    & 0.404 & 0.407 & 374.034 & 0.140 & 0.016 & 0.000 & 0.001 & 0.000 & 0.001 & 0.002 & 0.005 & 0.000 & 0.004 & 0.298 & 0.028 \\
\midrule
\multirow{2}{*}{\sc{\makecell{Policy\\extraction}}}
                    & 0.175 & 0.180 & 11.701 & 0.148 & 0.036 & 0.016 & 0.027 & 0.017 & 0.021 & 0.046 & 0.042 & 0.016 & 0.044 & 0.015 & 0.171 \\
                    & 0.193 & 0.184 & 18.702 & 0.172 & 0.038 & 0.049 & 0.027 & 0.017 & 0.021 & 0.048 & 0.044 & 0.017 & 0.046 & 2.584 & 0.185 \\
\bottomrule
\end{tabular}
}
\label{tab:res}
\end{table*}

\subsection{\kb Generation}
The results in Table~\ref{tab:res} show \review{promising} generalisation abilities from the LLM, demonstrated across 3 different domains. 

\subsubsection{Structure Building}
The LLM generated the correct \kb in \review{all the} five cases\review{, across varying numbers of pillars, blocks and sections.} The examples used for the few-shot prompt were similar to the tested \uc. \review{Notably, it correctly formulated the action to place the architrave on top of the pillars, an action that can be executed only at the end in Case 5.} 

\subsubsection{AGV}
The LLM was able to generate consistent \kbs from the examples. Only one error occurred in scenario 4, in which the LLM added an additional non instantiated predicate to the \textit{verify predicate} list. This is considered a minor error, since the Swipl interpreter immediately identified the error, and the correction consisted in removing such predicate from the list.

 \review{\subsubsection{Grippers} The LLM generated the correct \kb in three out of five cases, without any new example. The errors committed by GPT-4o were not conceptual errors, but rather simple syntax mistakes that could be easily detected and corrected by an expert. For instance, in both Case 4 and 5, the error consisted in using \texttt{ball3} instead of \texttt{ball3\_position}. Such errors could also be automatically identified and corrected by integrating syntax and and consistency-checking tools~\cite{duranti2024llm}.}

\review{Finally, given the nature of the observed errors, we also conducted additional tests using GPT-5-mini. This model exhibited improved generalisation capabilities and achieved a perfect score in the generation of the KB, confirming the robustness and scalability of our approach.}

\subsection{MDP and Policy Extraction}
The results of the generation of the MDP and for the extraction of the policy are shown in Table~\ref{tab:res}. 
For the considered examples, the generation of \kb and MDP took a negligible time, which allowed us to make multiple queries in a small time. This does not mean that the
framework could be used for online
generation in industrial scale applications.

Overall, performance varies significantly with the complexity of the MDP, i.e., the number of states and actions. The framework solves the tests for the AGV \uc--which includes between 35 and 189 states and 2 or 3 actions--consistently faster than the tests for the structure building \uc, which ranges from 17 to 1024 states and 16 or 27 actions. Notably, the time required to refine transition probabilities increases with the number of available actions per state. This factor has a substantial impact, often dominating the total computation time (Table~\ref{tab:res}).
Extracted policies respect the criteria written in the input queries both when maximising the probabilities of success or minimising the reward values. 

\begin{table}[t]
\centering
\caption{Simulation results. For the structure building \uc, we show the total number of actions and the number of actions that propose 2 or 3 equal blocks. For the AGV \uc, we compare the optimal policy with a faulty one in which there is a probability $p_f=0.4$ of choosing the wrong action.}
\label{tab:policyRes}
{\renewcommand{\arraystretch}{0.85}
\begin{tabular}{rcccccc}
\toprule
    &       & \multicolumn{3}{c}{\makecell{\sc{Structure}\ \sc{building}}} & \multicolumn{2}{c}{\sc{AGV}} \\
\midrule
    &       & \multicolumn{3}{c}{\sc{\# Actions}}  & \multicolumn{2}{c}{\sc{Success ratio}} \\ 
    & \sc{Test} &   \sc{  Tot   } &   \sc{   2    } &   \sc{   3   } & \sc{Opt}   & \sc{Faulty} \\
\midrule
\multirow{5}{*}{\rotatebox[origin=c]{90}{\sc{\makecell{Optimal\\policy}}}}
 &  1 &    90000 &    67754 &   22246 & 1.0000 & 0.8024 \\ 
 &  2 &    90000 &    67273 &   22727 & 1.0000 & 0.8591 \\ 
 &  3 &   150000 &   116596 &   33404 & 1.0000 & 0.8090 \\ 
 &  4 &   120000 &    20268 &       - & 0.0148 & 0.0163 \\ 
 &  5 &    70000 &    43454 &   16546 & 0.0697 & 0.0516 \\ 
\midrule

\multirow{5}{*}{\rotatebox[origin=c]{90}{\sc{\makecell{$\varepsilon$-greedy\\policy}}}} 
 &  1 &    90000 &    20000 &    3257 & 0.0380 & 0.0602 \\ 
 &  2 &    90000 &        0 &   24053 & 0.0003 & 0.0007 \\ 
 &  3 &   150000 &   113349 &       0 & 0.0126 & 0.0232 \\ 
 &  4 &   120000 &        0 &       - & 0.0126 & 0.0175 \\ 
 &  5 &    70000 &        0 &    1133 & 0.0166 & 0.0185 \\
\bottomrule
\end{tabular}
}
\end{table}

\subsection{Policy execution}
To provide a brief evaluation of policy behaviour, we tested the framework in \review{three} representative scenarios. 
In the structure building task, the probability-maximising policy \texttt{doneP} often used repeat block combinations (over 75\% duplicates in test 1), while the reward-oriented policy \texttt{doneR} created more diverse sets, demonstrating our approach's ability to bias policies toward reliability or diversity. In the AGV use case, the probability-maximising policy consistently achieved goals under ideal conditions (100\% success in tests 1–3) and performed well under faults (over 80\% success), though its performance decreased in tests 4–5 due to limited action effectiveness. Conversely, the reward-minimising policy struggled even without faults, with faults further reducing success, highlighting its fragility. These experiments show that our framework supports different optimisation objectives and exposes their trade-offs in robustness and diversity.

\subsubsection{Real-world experiment.}
\label{ssec:rw-experiment}
In the real-world experiment, we used a UR5e robotic arm, equipped with a 2-finger soft gripper (Figure~\ref{fig:lab-setting}), and involved a total of four subjects, including two authors and two additional participants, each of whom completed the test twice. In the first round, the user was instructed to build three pillars using blocks from the robotic arm. In the second round, they were informed of the building policy, i.e., building by layer. All participants successfully constructed the structure each time, confirming the correctness and robustness of the extracted policies. A video of the experiment can be found in the multimedia material.
\section{Discussion and Conclusions}
\label{sec:conlusion}

In this paper, we presented a complete framework that automates the generation of MDP policies from natural language queries. The proposed approach leverages state-of-art prompt-engineering techniques to construct a Prolog \kb, which is subsequently used to derive an MDP and its corresponding policy through \storm. The resulting policy can be conveniently stored in a state–action table, enabling straightforward deployment in real-world scenarios. Overall, this study provides a foundational step toward integrating natural language interfaces with formal decision-making frameworks, paving the way for more accessible and adaptive autonomous systems.

Despite its promising results, the framework has several limitations. The most critical limitation arises from LLMs inherent tendency to generate incorrect or inconsistent outputs. At present, the system does not include an automated mechanism for error detection or correction, and therefore primarily functions as a support tool for domain experts. Future work will focus on introducing a feedback loop for self-correction when the parser fails or after MDP analysis. Additionally, employing LLM ensembles or fine-tuned domain-specific models may help mitigate such inaccuracies.
A second limitation concerns the LLM’s inability to autonomously infer action probabilities, which currently requires expert intervention. To address this, we are investigating methods to estimate these probabilities from prior task demonstrations, such as video data or raw sensory inputs. Furthermore, while the framework is designed for expert users, its initial learning curve can be steep. We are developing comprehensive usage guidelines, prompt templates, and an intuitive user interface to improve accessibility and usability.
Another challenge involves the generation of few-shot examples to guide the LLM in creating effective knowledge bases. Although the model demonstrates strong generalisation capabilities across domains, curating high-quality examples remains essential for robust performance. To this end, we are preparing standardised templates and documentation to facilitate this process.
As future work, we also aim to extend the framework to support partially observable MDPs and hidden Markov models, thereby enhancing its ability to model uncertainty and improve interaction with human operators.

\bibliographystyle{IEEEtran}
\bibliography{biblio}



\end{document}